\documentclass[journal]{IEEEtran}
\usepackage{url}
\usepackage{float}
\ifCLASSINFOpdf
\usepackage[pdftex]{graphicx}
\DeclareGraphicsExtensions{.pdf,.jpeg,.png}
\else
\fi

\begin{document}

\textbf{This is a preprint. Please, cite us as:}
\\
A.L. Alfeo, M.G.C.A. Cimino, S.Egidi, B.Lepri, G. Vaglini, "A stigmergy-based analysis of city hotspots to discover trends and anomalies in urban transportation usage", IEEE Transactions on Intelligent Transportation Systems, IEEE, Vol. 19, Issue 7, Pages 2258-2267, 2018, (ISSN 1524-9050), doi: $10.1109/TITS.2018.2817558$.

\newpage
\title{A stigmergy-based analysis of city hotspots to discover trends and anomalies in urban transportation usage}

\author{Antonio L.~Alfeo, %~\IEEEmembership{Member,~IEEE,}
        Mario G. C. A.~Cimino, %~\IEEEmembership{Fellow,~OSA,}
        Sara~Egidi, %~\IEEEmembership{Fellow,~OSA,}
        Bruno~Lepri, %~\IEEEmembership{Fellow,~OSA,}
        and~Gigliola~Vaglini%~\IEEEmembership{Life~Fellow,~IEEE}% <-this % stops a space

\thanks{Antonio L. Alfeo is with the Department of Information Engineering, both at the University of Pisa, (largo Lucio Lazzarino 1, Pisa, Italy) and the University of Florence (Via di Santa Marta, 3, Florence, Italy). Email: luca.alfeo@ing.unipi.it,}% <-this % stops a space
\thanks{Mario G. C. A. Cimino, Sara Egidi, and Gigliola Vaglini are with the Department of Information Engineering, University of Pisa.   
Email: mario.cimino@unipi.it, s.egidi1@studenti.unipi.it, gigliola.vaglini@unipi.it.}% <-this % stops a space
\thanks{Bruno Lepri is with Bruno Kessler Foundation, via S. Croce, 77, Trento, Italy.  
Email: lepri@fbk.eu. }% <-this % stops a space
\thanks{Manuscript accepted on March 14, 2018}
}



\markboth{IEEE Transactions on Intelligent Transportation Systems (ACCEPTED PAPER - PREPRINT)}
{Shell \MakeLowercase{\textit{et al.}}}%: Bare Demo of IEEEtran.cls for IEEE Journals}


%\IEEEspecialpapernotice{(ACCEPTED PAPER - PREPRINT)}

\maketitle

\begin{abstract}
A key aspect of a sustainable urban transportation system is the effectiveness of transportation policies. To be effective, a policy has to consider a broad range of elements, such as pollution emission, traffic flow, and human mobility. Due to the complexity and variability of these elements in the urban area, to produce effective policies remains a very challenging task. 
With the introduction of the smart city paradigm, a widely available amount of data can be generated in the urban spaces. Such data can be a fundamental source of knowledge to improve policies because they can reflect the sustainability issues underlying the city.

In this context, we propose an approach to exploit urban positioning data based on stigmergy, a bio-inspired mechanism providing scalar and temporal aggregation of samples. By employing stigmergy, samples in proximity with each other are aggregated into a functional structure called trail. The trail summarizes relevant dynamics in data and allows matching them, providing a measure of their similarity. Moreover, this mechanism can be specialized to unfold specific dynamics.

Specifically, we identify high-density urban areas (i.e. hotspots), analyze their activity over time, and unfold anomalies. Moreover, by matching activity patterns, a continuous measure of the dissimilarity with respect to the typical activity pattern is provided. This measure can be used by policy makers to evaluate the effect of policies and change them dynamically. As a case study, we analyze taxi trip data gathered in Manhattan from 2013 to 2015.

\end{abstract}


% \fi
\IEEEpeerreviewmaketitle

\section{Introduction and Motivation}
\IEEEPARstart{S}{martness} 
and sustainability are two key aspects of the forthcoming transportation systems. Smartness provides transportation monitoring and control with qualities like real-time sensing and fast decision making. Sustainability aims to manage travel demand efficiently by means of environmentally friendly strategies, providing transportation systems with policies for long-term economic suitability \cite{haque:2013}. 
%\textcolor{blue}{
Specifically, a sustainable urban development demands adequate policy instruments aimed to handle and mitigate the increasing volume of traffic congestion, carbon emission, and air pollution. One of the most frequently used policy tools for the measurement and evaluation of transportation sustainability performance are indicators. Indicators can be defined as quantitative measures aimed to explain and communicate complex phenomena simply, including trends and progress over time \cite{eea:2005}. In order to provide effective measures of sustainability of transportation activities, it is essential to define indicators’ purpose and scope. During the last two decades, a number of international initiatives addressed the development of indicators aimed to achieve a more sustainable transportation on the local, regional, and global levels, by involving both scientific community and policy-makers \cite{indicator2}. 
However, as of today, there is no standard or common agreement about the set of indicators to be used to assess transportation sustainability. Many works in the field perform an impact-based classification by employing a three-dimensional framework based on economic, environmental, and social impacts \cite{dobransky:2007}. The proposed indicators are in general calculated by exploiting commonly available data sources  \cite{indicator2}.
Thanks to the pervasive technology supporting the smart city strategy, some of these indicators may be calculated via big data fed by on-board or fixed sensors. As an example, data obtained by smart cards \cite{park:2008} can be used to describe the characteristics of public transit usage, such as the number of trips for different transit modes, and travel time distribution for all transit modes and user types. Another example could be the GPS-enabled vehicles, which can provide a more comprehensive view of the factors shaping transportation emissions and efficiency, by analyzing passenger occupancy and trip density by location and time \cite{an:2011}. Finally, air quality monitoring systems can be used to monitor local pollution emission(\cite{matte:2013},\cite{zheng:2013}) and noise emission \cite{zheng:2014}. These sources allow enhancing the precision of the investigation, providing insights about sustainability issues on specific urban locations and the moment in time. 
According to \cite{indicator1}, \cite{indicator2}, \cite{indicator3}, \cite{indicator4}, some of the widely accepted indicators assessing the sustainability of transportations are derivable from localized sensors data. Below we present a list of indicators arranged according to their addressed purpose.
\begin{enumerate}
\item Transportation system efficiency:
\begin{enumerate}
\item Total vehicle-mileage traveled by motorized (private and public) and non-motorized (bikes and pedestrians) traffic participants;
\item Portion of travels made by private car;
\item Total vehicle-mileage traveled in urban-peak conditions;
\item Delay per trip;
\item Vehicle occupancy per travel (as passenger-vehicle);
\item Frequency and mean duration of traffic congestions;
\end{enumerate}
\item Pollutions levels:
\begin{enumerate}
\item Total emissions of greenhouse gases (e.g. CO\textsubscript{2}); 
\item Total emissions of air pollutants (e.g. PM\textsubscript{10});
\item Total emissions of noise pollutants;
\end{enumerate}
\item Land use: 
\begin{enumerate}
\item Percentage of population living in proximity to public transportation facilities (e.g. transit station); 
\item Total land area consumed by transportation infrastructure;
\item Total land area consumed by cars;
\end{enumerate} 
\end{enumerate}
%}
%Sustainable indicators are usually exploited by means of conventional statistical approaches, which are commonly used by public and private organizations to evaluate their overall activities performances. However, 
According to the moment in time and the location observed, an indicator may show different patterns. The detection of significant patterns usually involves a complex system modeling, due to largeness and complexity of data underlying the indicator. Thus there is the need for a data mining algorithm aimed to provide pattern identification, detection of behavioral regularities, and comparison between different traffic phenomena \cite{prabha:2016}. 

In this paper, we propose a novel approach to analyze urban positioning data. Both complexity and largeness of data can be handled by employing computational techniques belonging to the \emph{emergent paradigm}. Emergent paradigms allow avoiding the explicit modeling of dynamics, which 
%can be very difficult to make due to dynamics complexity and 
works only under the assumption formulated by the designer. Emergent paradigms, instead, offer model-free computational approaches, characterized by adaptation, autonomy, and self-organization of data \cite{vernon:2007}. 

In particular, the emergent paradigm provides a biologically inspired aggregation mechanism known as \emph{stigmergy}. By using stigmergy each sample position is transformed into a digital pheromone deposit. Deposits accumulate with each other according to their proximity while progressively evaporating. This simple mechanism intrinsically embodies the time domain, thus it is used to unfold significant spatiotemporal dynamics in the data \cite{alfeo:2016}.

Among all the available sources of movement data for smart city application, location aware vehicles provide us the opportunity to help urban planners and policy makers in mitigating traffic, planning for public services and resources, and properly manage infrequent events \cite{liu:2012}. 
%Post-Review1
%\textcolor{blue}{
However, both public transportation and private vehicles provide quite predictable GPS traces, because they are due to predetermined routes or personal routines (i.e. to and from work). On the other hand, %}
%Post-Review1
GPS-enabled taxis, represent both a transit-complementary door-to-door transportation mode and a source of real-time human mobility information \cite{veloso:2011}. Indeed, taxicabs play a prominent role as a transportation mode in metropolitan areas, e.g., in New York City, over 100 companies operate more than 13,000 taxicabs with a daily demand of 660,000 passengers \cite{zhang:2013}. 
%Post-review2
%\textcolor{blue}{
Moreover, by continuously serving a wide diversity of passengers in the city, taxis GPS traces can provide a detailed glimpse into motivation and characterization of population’s urban mobility. %}
%Post-review2
However, regular taxicab services becomes inefficient during urban-peak conditions, e.g., extreme weather or special events (\cite{xianyuan:2014}, \cite{neuwirth}), producing unnecessary traffic, pollution, energy consumption, and causing the increase of passenger's waiting time \cite{urbComp}.
Thus, further investigation aimed at analyzing taxi-based transportation system is needed.
%PostReview3 Castro & Dataset
%\textcolor{blue}{
In this context, an exhaustive survey of taxis' trip data analysis is provided in \cite{castro:2013}. According to \textit{Castro et al}, existing works can be divided into 3 categories: 
(i) \textit{Social Dynamics}, which aims to analyze the collective behavior of urban population, detect the most visited areas in the city (hotspots), characterize their functionalities, and study the mutual relationships among different urban areas; the purpose of this kind of analysis is to determine the effectiveness of transportation systems and to provide guidance for needed improvements; 
(ii) \textit{Traffic Dynamics}, i.e. the analysis of the traffic flow through the city road network, providing results about congestion durations and levels in a given city area; congestions have a remarkable impact over the travel time and on the occurrence of adverse events (e.g., accidents), and can be used to estimate pollution levels \cite{guhnumann:2004}; 
and (iii) \textit{Operational Dynamics}, i.e. the analysis of taxi’s trajectory aimed to investigate taxi drivers’ behavior, providing route planning insights for unfolding anomalies in urban mobility.
According to the authors in \cite{castro:2013}, many works in the field deal with more than one of the proposed categories in their analysis, and this is the case of our work.
Specifically, we exploit taxis' trip data provided by Taxi and Limousine Commission (TLC) of New York City. All taxis of NYC are equipped with FCD (floating car data) devices, which manage localization and card payments data, and enable taxicab drivers and passengers to receive information from the Taxi and Limousine Commission \cite{TPEP}. FCD records include pick-up and drop-off positions, timestamp, and number of passengers, which feed the Taxi Trip Origin-Destination (OD) dataset.% }
%PostReview3
We analyze it in order to uncover city hotspots, characterize human mobility patterns, and detect anomalous occurrences by using an approach based on stigmergy.

The paper is organized as follows. In section 2 we discuss the related works. In section 3 we present our approach.  We detail data preprocessing and experimental setup in section 4. In section 5 results obtained by analyzing taxi traces gathered in NYC during the years 2013, 2014 and 2015 are shown. Finally,  we conclude our work and discuss future avenues of research in section 6.



\section{Related Work}
Recently, the wide availability of taxi trip data has produced a significant number of works aimed to mine urban dynamics by exploiting this kind of data. 

%With the aim of unfolding patterns in positioning data, authors in \cite{farrahi:2008} and \cite{farrahi:2011} use a generative statistical model, called Latent Dirichlet Allocation (LDA), to automatically discover location-based daily routine patterns by using mobile phones GPS data. 
In \cite{peng:2012} the authors use non-negative matrix factorization (NMF) algorithm to decompose taxi activity levels and extract three basic patterns. Those patterns represent respectively: (i) commuting between home and workplace, (ii) business traveling between two workplaces, and (iii) leisure trips from or to other places. Furthermore, authors model the relative daily deviation of the traffic flow in each category. 
In \cite{zhang:2015} authors analyze taxi traces in order to model the typical pattern of passenger flow in an urban area; by applying this model authors were able to compute the probability that an event happened, and  measure the impact of the event by analyzing anomalous patterns in passenger flow via Discrete Fourier Transform. 
An Interactive Voting-based Map Matching Algorithm is used in \cite{pan:2013} to map GPS trajectories. This mapping is aimed to characterize typical drivers’ behaviors and discover abnormal ones. Finally, the authors mine the cause of the anomaly by checking data gathered by social networks. 

One of the main issues concerning the analysis of this kind of data is their dimensionality. 
Many approaches handle it by focusing on specific areas (hotspots) whose high concentration of events or samples can summarize the most relevant dynamics occurring in data \cite{hu:2014}, \cite{alfeo:2017}.

In the literature, urban hotspots are typically divided in two categories: (i) regular and (ii)
occasional. Areas comprising many points of interest such as movie theaters, commercial buildings, hospitals, schools, colleges, etc. are prime examples of regular hotspots. Occasional hotspots are those areas where any incident has taken place. An incident is defined as an unexpected event that temporarily disrupts the mobility flow, e.g., car crash, marathon, VIP passing area, etc. However, the most of the studies firstly consider regular hotspots. Li. et al \cite{li:2012} proposed and developed an improved auto-regressive integrated moving average (ARIMA) for detecting urban mobility hotspots using taxi GPS traces; moreover, the patterns of pick-ups occurring in these city locations are extracted and analyzed. 
Other works, such as the one from Makrai \cite{makrai}, provide a statistical approach for the detection of hotspots in New York City by means of a distributed environment. 
Authors in \cite{keler:2016} use OPTICS in order to exploit taxi drop-off positions, extracting hotspots from density-connected point clusters. Cluster results are then assigned as daily taxi drop-off hotspots. 
Recently, Lu et al. \cite{lu:2016} developed a monitoring system performing spatiotemporal analysis on taxi trip data to find seasonal hotspots. This result is achieved by using DBSCAN clustering algorithm with pick up and drop-off locations every fixed amount of time.

\section{Approach description}
Our idea is to represent taxi traces as deposits of pheromones, akin to the ones ants release when exploring and searching for food. Isolated pheromones progressively evaporate and disappear, whereas pheromones which are subsequently  deposited in proximity with each other aggregate as a trail. The trail guides the ants
while seeking for food, providing them an effective self-organization mechanism, known as stigmergy \cite{theraulaz:1999}. 
The emulation of this mechanism in the context of data processing enables the unfolding of spatial and 
temporal dynamics in data \cite{alfeo:2016} by providing information self-organization \cite{vernon:2007}. Specifically, the principle of stigmergy is used (i) to discover a set of locations (hotspots) characterized by the highest spatiotemporal density in data; (ii) to unfold hotspot activity patterns, (iii) to match those patterns in order to detect anomalies in hotspot activity. The overall processing schema has been developed and integrated with the MATLAB framework.

\subsection{Hotspot Discovery}
\begin{figure}[!t]
\centering
\includegraphics[width=2.3in]{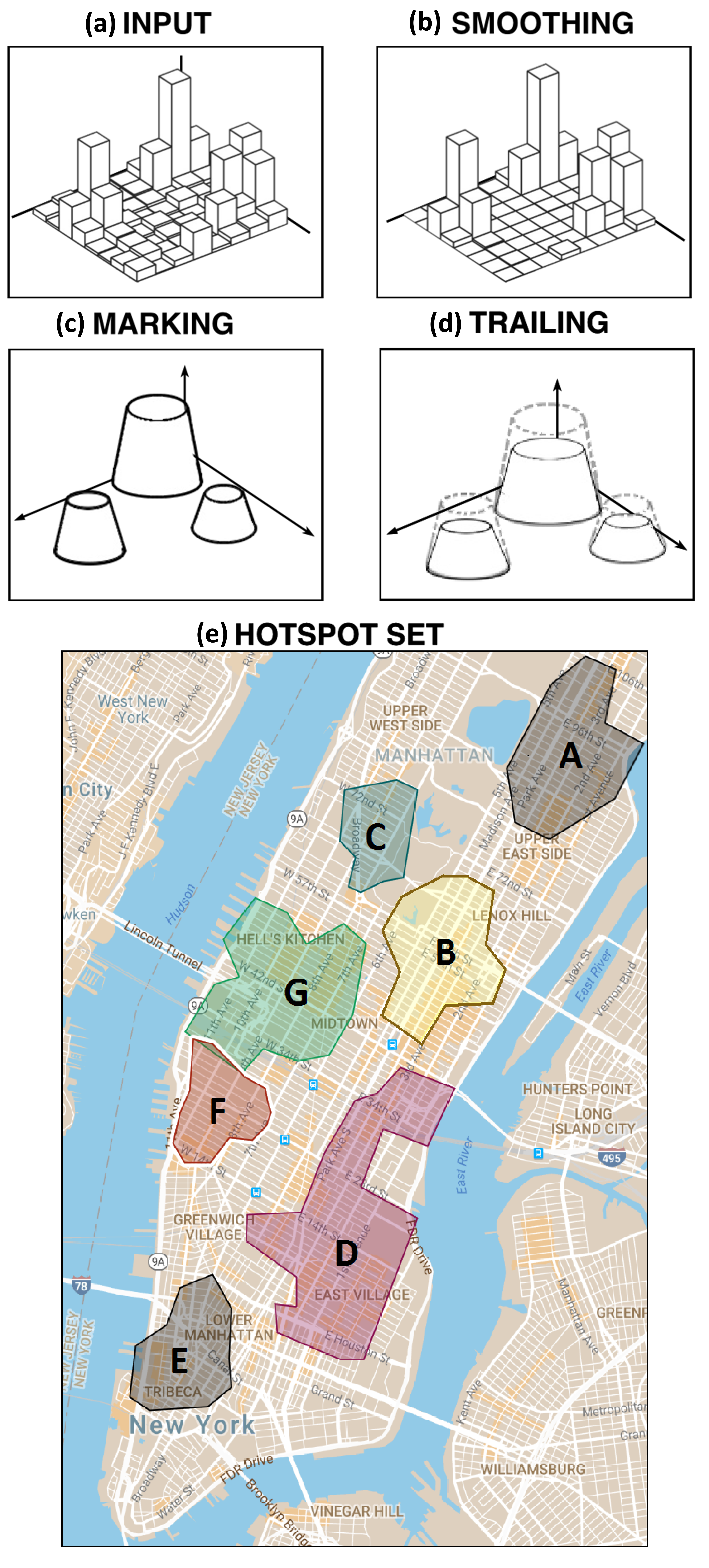}
\caption{The overall process of hotspots discovery in Manhattan.}
\label{img:HD}
\end{figure}

With the aim of discovering locations characterized by high density of taxi passengers activity (activity, for short), we exploit the number of people being picked up or dropped off in a given time slot and in a given location.
During each step of the analysis, samples corresponding to a given time slot are provided to the system and processed by a four-stages procedure. At the beginning, the Smoothing process
removes irrelevant samples' values and highlights significant ones (Fig.\ref{img:HD}b), by treating them with a sigmoidal function. The marking process releases a mark in a tridimensional spatial environment in correspondence of each smoothed sample position (Fig.\ref{img:HD}c). Each mark is defined by a truncated cone with a given width and an intensity (height) equal to the sample value. The marks aggregate form the trail, whose intensity is subject to evaporation, i.e., the trail intensity is decreased
by a constant value $\delta$ at each step of the analysis (Fig.\ref{img:HD}d). Eq. \ref{eq:trail} describes the trail $T$ at time instant $i$.

\begin{equation}
\label{eq:trail}
T_{i} = (T_{i-1} - \delta) + Mark_{i}
\end{equation}

Due to aggregation and evaporation provided by the trailing process, sparse marks progressively disappear. On the contrary,
subsequent deposits in proximity with each other counteract the evaporation producing a stable
trail, which highlights spatiotemporal density in data.

The hotspots are determined as the overlap of the city areas corresponding to the most relevant trails, obtained by analyzing data corresponding to early morning (i.e., 3a.m.-8a.m.), morning (i.e., 9am-2pm), 
afternoon/evening (i.e., 3pm-8pm), and night (i.e., 9pm-2am) time slots. Each hotspot is represented by a 
set of coordinates that bound a city area inside a polygon.
The hotspots identified in Manhattan (New York City) are shown in Fig. \ref{img:HD}e. Their locations correspond to East
Harlem - Upper East Side (A), Midtown East (B), Broadway (C), East Village - Gramercy - MurrayHill (D), Soho - Tribeca (E), Chelsea (F) and Time Square
- Midtown West - Garment (G).

\subsection{Stigmergic Receptive Field}
\begin{figure*}[!t]
\centering
\includegraphics[width=6in]{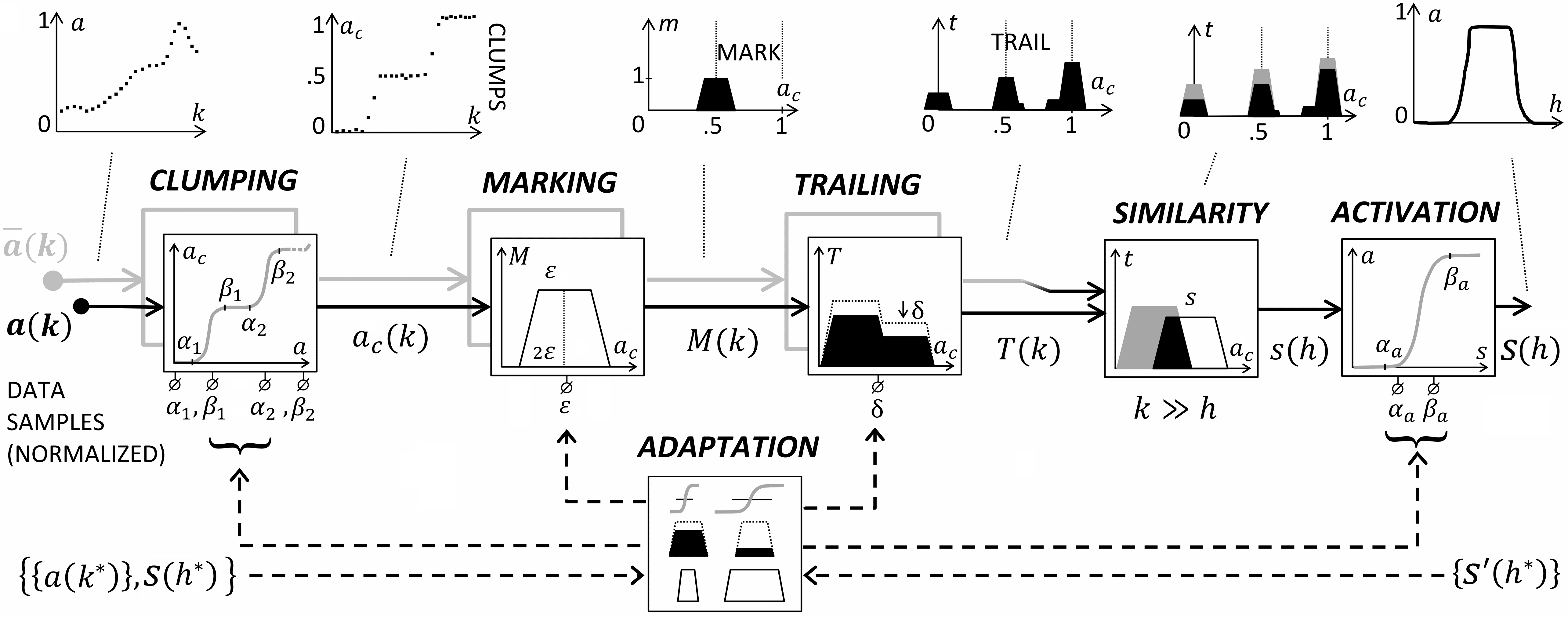}
\caption{The architecture of a stigmergic receptive field.}
\label{img:SRF}
\end{figure*}

The result of the hotspot identification is a set of urban areas in which the most relevant activity dynamics occur. For each of them, we extract the activity time series by gathering the amount of activity occurred in the hotspots during the day.
The hotspot activity time series are analyzed by means of the Stigmergic Receptive Field (SRF), a stigmergy-based processing schema which provides a similarity measure of a given couple of time series ($a(k)$ and $\bar{a}(k)$ in Fig. \ref{img:SRF}).

The overall architecture of the SRF is composed by a five-stages processing pipeline, shown in Fig. \ref{img:SRF}.

Input samples are firstly treated by the Clumping process. It is aimed to reduce microfluctuation in data while highlighting the dynamics occurring among relevant information levels. In our application, 3 levels are used to specify the activity behaviors, namely Low, Medium and High. The Clumping process is implemented by treating each input sample with a double sigmoid function, which is parametrized by means of its inflection points ($\alpha_{1},\beta_{1},\alpha_{2},\beta_{2}$).
In correspondence of each clumped sample, a mark is released in a bidimensional virtual environment by the Marking process. The mark is defined as a trapezoid with given intensity (i.e., height) and width ($\epsilon$).
The Trailing process handles the aggregation of the marks in the trail, and the trail temporal decay, i.e., the decrease of its intensity of a given amount $\delta$ at each step of time.
The trail can be considered as a short term memory summarization of the spatiotemporal dynamism occurring in data.

Both time series ($a(k), \bar{a}(k)$) provided to the SRF undergo these processing stages independently, until the Similarity process compares their trails ($T_1,T_2$) using the Jaccard coefficient \cite{jaccard:2013} as defined in Eq. \ref{eq:jaccard}.

\begin{equation}
\label{eq:jaccard}
S = \frac{|T_{1} \cap T_{2}|}{|T_{1} \cup T_{2}|}
\end{equation}

The provided similarity values are between 0 
(completely different trails) and 1 (identical trails). Finally, the Activation process transforms the similarity values by treating them with a sigmoidal function (Eq. \ref{eq:sigm}). Specifically, activation process lowers insignificant similarity values and enhances relevant ones depending on the values of the sigmoid inflection points ($\alpha_{a},\beta_{a}$).  

\begin{equation}
\label{eq:sigm}
f(x,\alpha_{a},\beta_{a}) = \frac{1}{(1+e^{-\alpha_{a}(x-\beta_{a})})}
\end{equation}

In order to have an effective sample processing the SRF should be properly parameterized.
For example, low trail evaporation causes early activation, whereas high trail evaporation generates a trail consisting of the latest marks only, and preventing pattern reinforcement. 
Specifically, the SRF parameters are: (i) the clumping inflection points $\alpha_{c1},\beta_{c1},\alpha_{c2},\beta_{c2}$; (ii) the
mark width $\epsilon$; (iii) the trail evaporation $\delta$; and (iv) the activation inflection points $\alpha_{a},\beta_{a}$.
The SRF parameters are adjusted by the Adaptation process. It uses the Differential Evolution (DE) algorithm \cite{alfeo:2017}, in order to minimize Mean Square Error (MSE, Eq. \ref{eq:mse}), which is computed as the difference between desired ($S'(h^*)$) and actual ($S(h^*)$) output value on a training set of N labeled couples of activity time series ($\left\{a(k^*),S'(h^*)\right\}$).

\begin{equation}
\label{eq:mse}
Fitness = \frac{\sum_{i=1}^{N} (|S (i^*) -S'(i^*)|^2)}{N}
\end{equation}

\subsection{Stigmergic Perceptron}
\begin{figure}[!t]
\centering
\includegraphics[width=2.5in]{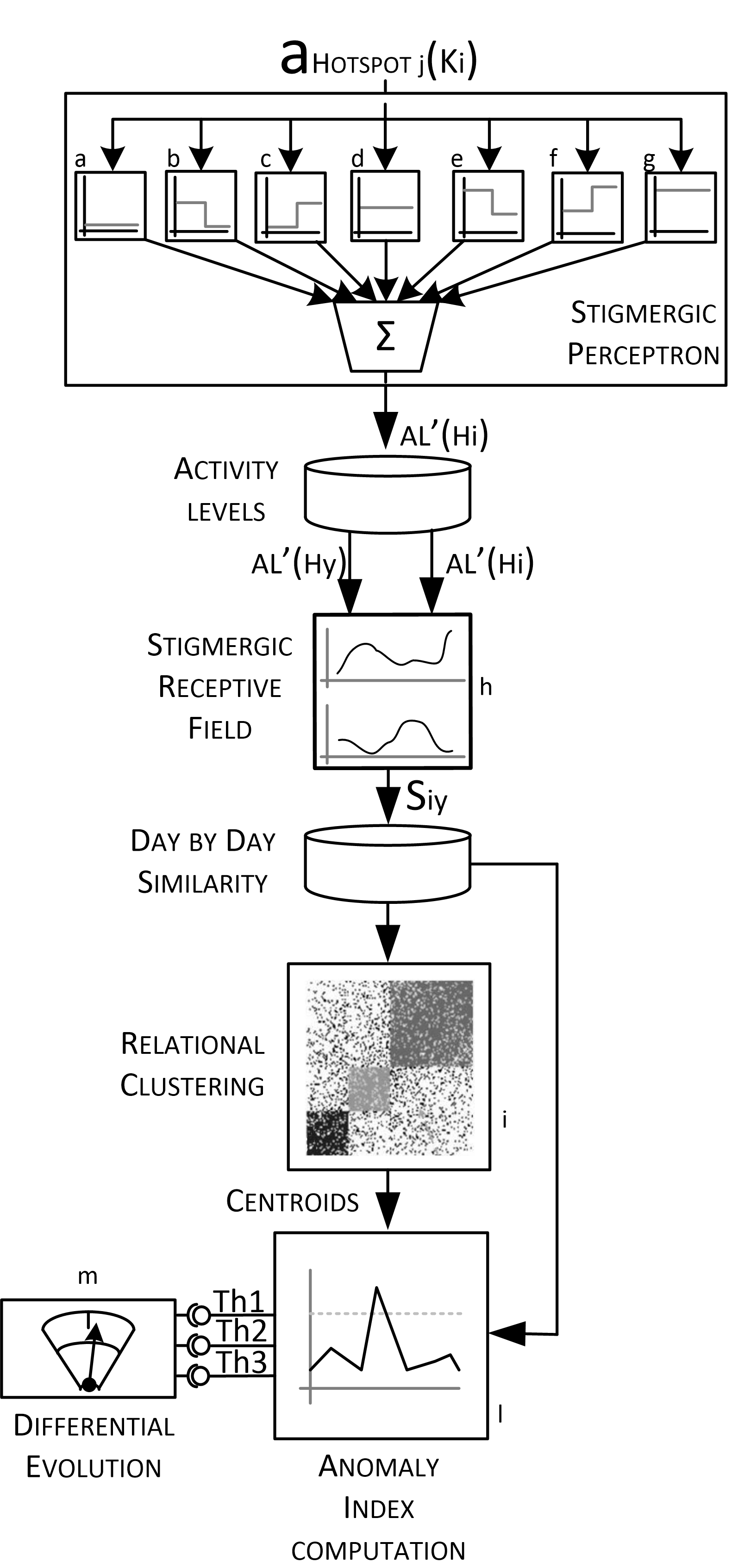}
\caption{The overall processing of the activity samples.}
\label{img:ARCH}
\end{figure}

Let us suppose to have a pure form time series which embodies a behavioral class (i.e., an archetype).
The SRF can detect this specific behavior in the actual time series, by processing it together with the archetype. An example of behavioral class in our domain is “Rush-Hour” (Fig. \ref{img:ARCH}a), which correspond to the behavior of the activity occurring in the hotspot when people movement is at its highest rate. Other classes provided are Asleep (Fig. \ref{img:ARCH}g), i.e., the
hotspot at its lowest activity level; Falling (Fig. \ref{img:ARCH}f), i.e., the transition between regular activity
and its calm down; Awakening (Fig. \ref{img:ARCH}e), i.e., the waking up of urban activity following a calm phase; 
Flow (Fig. \ref{img:ARCH}d), i.e., the hotspot at its operating capacity; Chill (Fig. \ref{img:ARCH}c), i.e., the calm down of the 
hotspot activity after a rush hour; Rise (Fig. \ref{img:ARCH}b), i.e., the transition to the most intense activity level.

A set of SRFs aimed to recognize these archetypes can be arranged into a connectionist topology, obtaining a Stigmergic Perceptron \cite{avvenuti:2016}. By forming a linear combination of the SRFs outcomes, the Stigmergic Perceptron provides an assessment of the current behavior of the input time series among all the classes provided. 
Specifically, the output of the SP is calculated as the average of the SRFs enumerations (represented as
1-to-7 in our application case) weighted by their output similarities (Eq. \ref{eq:activity}). 
The SP output is called activity level and is defined between 0 and N (i.e., the number of archetypes).

\begin{equation}
\label{eq:activity}
ActivityLevel = \frac{\sum_{i=1}^{N} (S(i) * i)}{\sum_{i=1}^{N} (S(i))} 
\end{equation}

\begin{figure}[!t]
\centering
\includegraphics[width=2.8in]{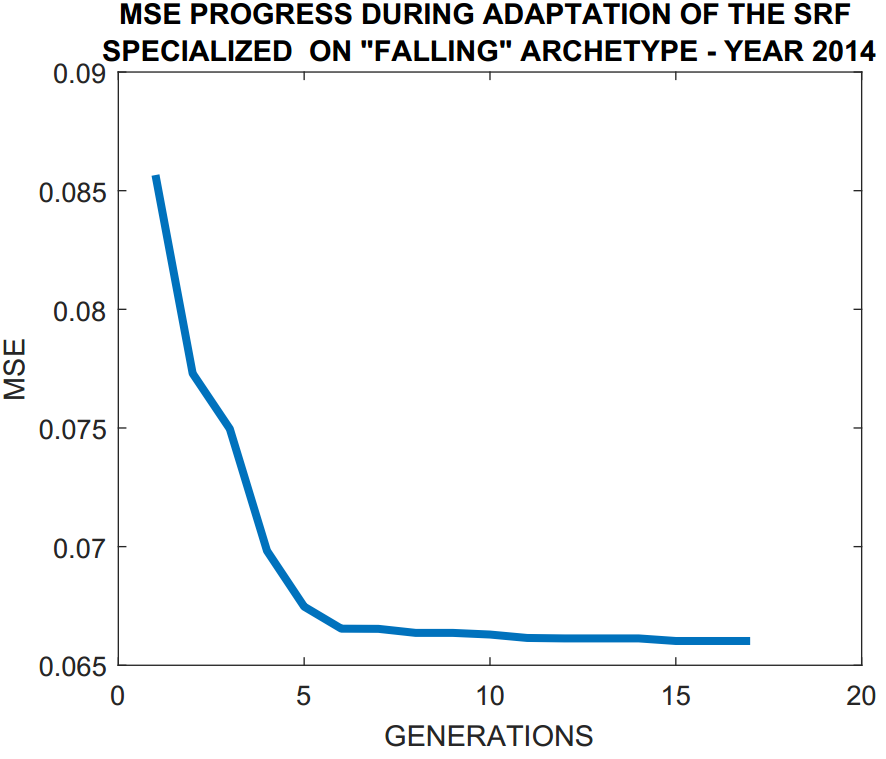}
\caption{The progressive decrease of the error in the assessment of the "Falling" archetypes (in year2014), during adaptation process.}
\label{img:DE}
\end{figure}

In order to prevent multiple activations of SRFs in the same SP, their Adaptation process is a two phases 
procedure: (i) the Global Training phase is aimed to provide a suitable interval for each SRFs parameters according to their sensitivity; specifically, the interval for the evaporation rate, that is the most sensitive SRF parameter, is determined considering the narrowest interval including the
fitness values above the 90th percentile, while the intervals for the other parameters can
be provided according to the application domain constraints; 
(ii) the Local Training phase is aimed to find the optimal values for every parameter and each SRF
by exploiting their Adaptation process and the interval determined in the Global phase; the training set 
for each SRF is made by half signals belonging to its behavioral class, and half belonging to the behavioral classes of adjacent SRFs. As an example, the progressive decrease of the MSE provided by this adaptation phase is shown in Fig. \ref{img:DE}.

A properly trained SP produces a time series of activity levels by transforming a given time series of activity samples. The activity levels time series can be considered as a higher level characterization of the hotspot activity during an entire day.
The source code of our Stigmergic Perceptron has been publicly released on the MATLAB Central File Exchange \cite{sourceCode}.

\subsection{Anomaly Degree Computation}
In order to detect anomalous activity level patterns, we employ a further SRF
%. We feed this SRF with the SP outcomes, arranging them in a multilayer topology aimed to handle big data complexity by providing information abstraction. Therefore, this SRF will provide a
aimed to measure of the similarity between two activity levels time series gathered in different days (Fig. \ref{img:ARCH}h).

\begin{figure}[!t]
\centering
\includegraphics[width=3.5in]{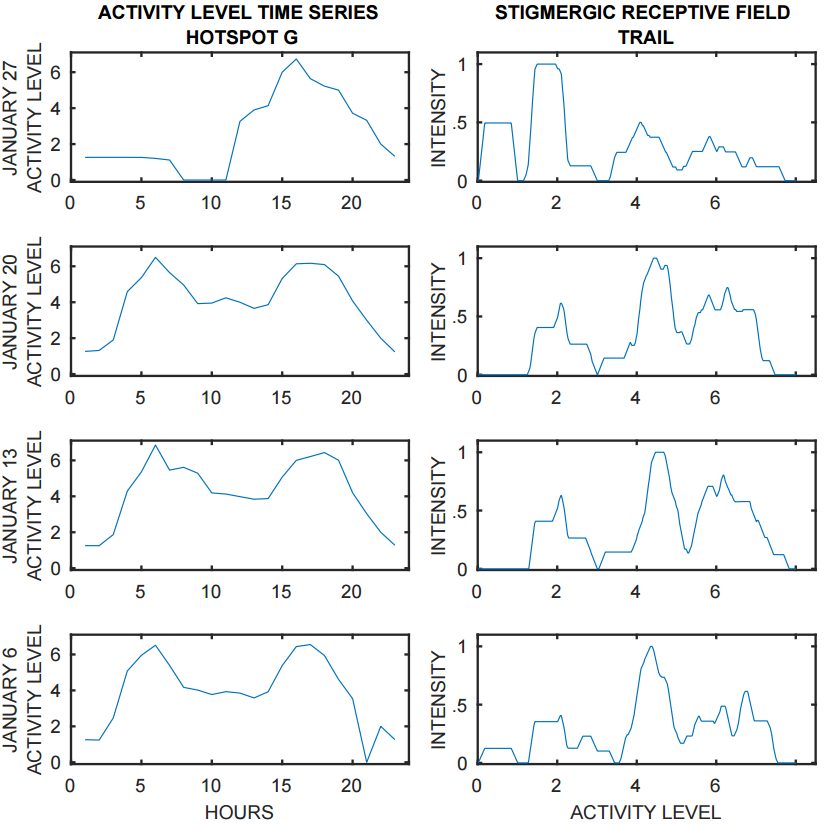}
\caption{Examples of trails generated by the activity level time series gathered in different working days. Jan-27 seems to be characterized by an anomalous pattern.}
\label{img:TRAILS}
\end{figure}

The Adaptation process of this SRF aims to minimize the difference (in terms of MSE) between computed and ideal similarity over a training set, by tuning mark width $\epsilon$, trail evaporation $\delta$, and
activation thresholds $\alpha_{a}, \beta_{a}$. The training set is composed by different (i.e., 900) couples of activity level time series whose similarity is supposed to be 1, if they belong to the same activity level behavioral class, 0 otherwise. As an example, an activity level behavioral class can be ``Working-Day" which is a day whose activity is mainly affected by working routines. Other activity level behavioral classes provided in our analysis are ``Entertainment-Days" (usually occurring in Fridays and Saturdays) and ``Leisure-Days" (usually occurring in Sundays).

As an example, Fig. \ref{img:TRAILS} shows the activity level obtained by analyzing activity in the hotspot G in every Tuesday of January 2015. The corresponding stigmergic trails are shown beside them. Here, an activity level time series differs from the other working days shown. Indeed, January 27 is characterized by an anomaly, due to winter storm 'Juno' that was hitting the city during that day.
 
\begin{figure}[!t]
\centering
\includegraphics[width=2.9in]{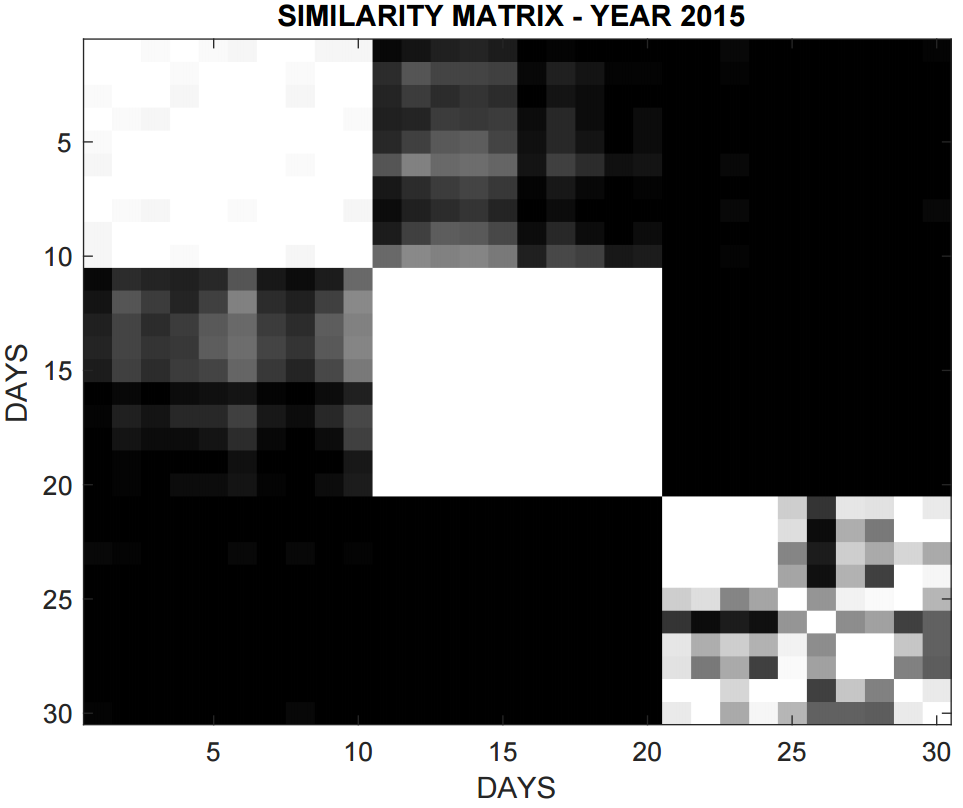}
\caption{Similarity matrix obtained by analyzing patterns gathered during year 2015.}
\label{img:SIM}
\end{figure}

By exploiting the SRF similarity measure we can match all the activity level time series of the SRF's training set and store their similarity values into a similarity matrix. As an example, in Fig. \ref{img:SIM}, we show the similarity matrix obtained by analyzing patterns gathered during the year 2015. Patterns are arranged by behavioral class, i.e., Working-Day (days 1-10), Entertainment-Day (days 11-20), and Leisure-Day (days 21-30). Here, the similarity value obtained by matching two patterns gathered in different days is represented by the color of the corresponding box. The whitest the box, the higher the similarity. As expected, the similarity values appear to be higher only with couples of days belonging to the same activity level behavioral class.

This similarity matrix is processed by a relational clustering technique (Fig. \ref{img:ARCH}i) in order to group similar daily activity levels. 
Specifically, we employ Fuzzy C-Means using as number of clusters the number of daily activity behaviors taken into account in the analysis (i.e. 3). The fuzzy clustering generates, for each activity level time series, a membership degree for each behavioral class. We can determine which are the most representative days for each behavioral class by selecting the activity level time series which are closest to the corresponding cluster centroid in terms of euclidean distance.

An activity level time series obtained by a typical day is expected to exhibit high similarity with respect to the activity level time series obtained by the most representative days of its behavioral class. Thus, we compute the Anomaly Index of current day $d$ by exploiting the average of its similarity $S(d,i)$ with respect to its most representative N (i.e., 5) days, as detailed in Eq. \ref{eq:AI}. The Anomaly Index is defined between 0 (typical daily behavior) and 1 (very anomalous daily behavior).

\begin{equation}
\label{eq:AI}
AnomalyIndex(d) = | \frac{\sum_{i=1}^{N} (S (d, i))}{N}  - 1 |
\end{equation}

\begin{figure}[!t]
\centering
\includegraphics[width=3.3in]{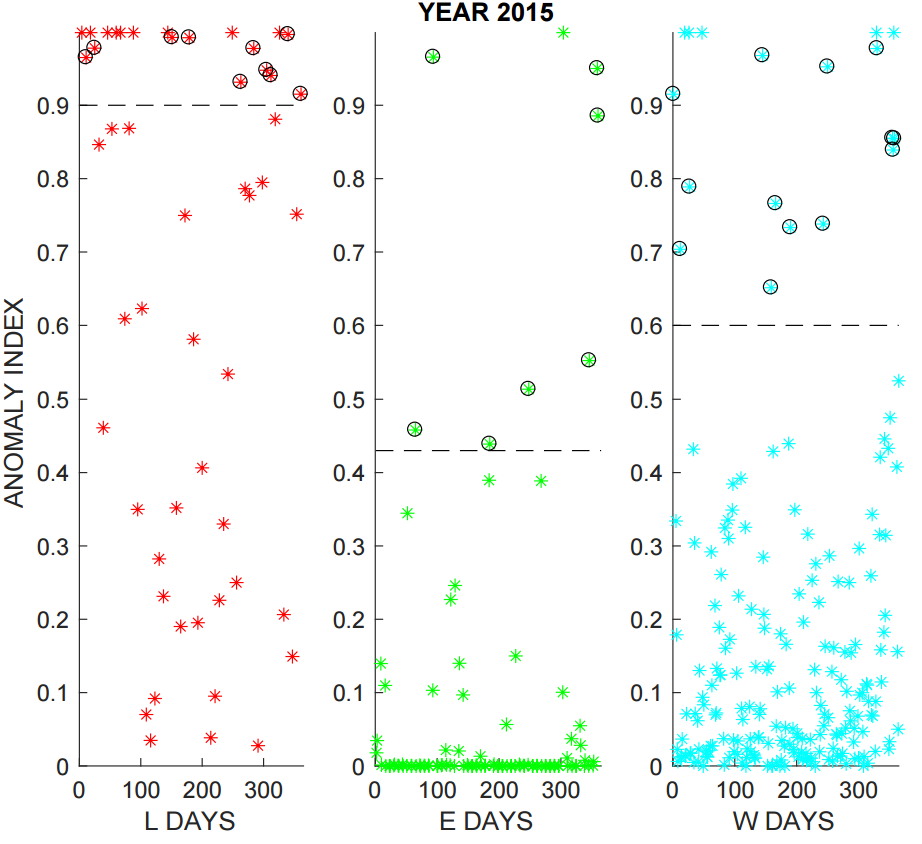}
\caption{Classification of anomalies (circle) and typical days (stars) obtained analyzing data gathered during 2015.The thresholds are reported in dashed line.}
\label{img:2015A}
\end{figure}

In order to discern typical days from anomalies, an Anomaly Index threshold for each activity level behavioral class must be defined.
These thresholds have been determined by using DE (Fig. \ref{img:ARCH}m) in order to minimize the classification error (i.e., the percentage of correctly classified days) over all the days of the year under analysis, given a set of known anomalies.
As an example, Fig. \ref{img:2015A} shows the classification of each day gathered during the year 2015. Known anomalies are characterized by a circle. The Anomaly Index thresholds are in dashed line.

\section{Experimental Setup}
We analyze data provided by the Taxi and Limousine Commission of New York City, containing details about all taxi trips occurred during 2013, 2014 and 2015 in Manhattan \cite{dataset}. 
%Post-review5
Each trip is reported with its taxi ID, number of passengers, together with latitude, longitude, and time-stamp of pick-up and drop-off.
%post-Review5
Data have been pre-processed in order to (i) remove missing values and (ii) discretize data in spatiotemporal buckets characterized by length and width of 10 foot, and duration of 5 minutes.

For the hotspots investigation, the period under analysis comprised both February
and June. This period has been chosen since it can capture different seasonal behavior without being influenced by the presence of many holidays. 

The activity time series extracted for each observed day has been normalized by using the min-max procedure. 

Both global and local training phases of the SP are provided with a training set generated 
by applying random spatial noise and temporal shift to the pure archetype time series.
The SP training set is composed of 70 time series (10 for each SRF), and the expected similarity is
1 if the current time series has been generated by the archetype on which current SRF must be specialized, 
0 otherwise. 

The SP outcome is processed by a further SRF, which is trained to measure activity level time series similarity, according to the daily behavioral classes provided, namely: (i) Working days which are expected to
fall between Monday and Tuesday, when commuters and working routines deeply affects the crowd movements; 
(ii) Entertainment days, which are expected to fall on Friday and Saturday, and are characterized by high nocturnal activity due to the nightlife; (iii) Leisure days, which are expected to fall on Sunday, and are 
characterized by minor transportation usage. The training set is composed of 30 activity level time series, i.e., 10 time series representing the typical patterns of each behavioral class. The target similarity of each possible match is 1 if time series falls in the same behavioral class, 0 otherwise. 

The typical pattern of each behavioral class has been determined by performing an exploratory analysis of the activity time series according to the following features:
(i) A label (low, medium, high) describing the range of the values contained in the initial part of the time series;
(ii) A label (low, medium, high) describing the range of the values contained in the final part of the time series;
(iii) The instant in which the mean value of the whole series is reached (initial and final part are not considered);
(iv) The instant in which the maximum value of the whole series is reached (initial and final part are not considered);
(v) The duration of the time series at its higher level.
We select different activity series for each behavioral class (Working, Entertainment, and Leisure) in order to extract these features. A suitable range of values for each feature and each behavioral class is provided by choosing the minimum and the maximum among the values obtained. 
By means of this features range, we can describe the overall typical pattern of each behavioral class. Moreover, the time series whose features exceed the features range of its expected behavioral class, can be annotated as an anomaly. As an example, the features of an activity time series gathered during a Thursday, are expected to fit the feature ranges of Working-Day behavioral class. If it does not occur the time series is annotated as an anomaly. 

%Specifically, Working days can be characterized by low initial and final mean values of the activity time series; moreover, they reach suddenly both mean and maximum activity earlier with respect to other behavioral class, due to working routine. The maximum activity duration is longer than the one exhibited by other behavioral classes.
%In Entertainment days, activity level reaches both its mean and maximum values later with respect to working days; moreover, due to night life occurring in this class of days, the activity time series remains high till the end of the series and the mean value in its initial are close to the mean value of the overall series.
%In Leisure days, the initial part of the activity time series shows the highest activity level occurring in leisure day, due to the end of the nightlife (Leisure days usually follows Entertainment days); indeed the rest of the activity level time series shows smooth transition to its mean value, a brief transition to a higher activity level (which is not the maximum of the series) and ends with low mean value, just like the Working days (which usually follows Leisure days).

\section{Results}
Depending on the land usage of the city area underlying each hotspot, some daily activity behaviors may not emerge. As an example, the Entertainment-Day behavior is mainly caused by the presence of clubs or other entertainment-oriented business that may attract the nightlife. Thus, the hotspots underlying a mixed usage zones are the most promising ones for the analysis, since our aim is to characterize all the aspects of the city life. 
%\textcolor{blue}{
According to official land use (publicy available at \cite{zola}), each city block can be classified into the following categories: commercial, residential, industrial, transportation space, institutional, open/recreational space, parking or vacant. By considering the distribution of these categories in each hotspot it can be evaluated how diversified the usage of that area is, and therefore the related amount of the mobility dynamics. Specifically (i) Hotspot A is primarily residential and secondly institutional; (ii) Hotspot B is mainly residential and commercial; (iii) Hotspot C is principally open space and residential; (iv) Hotspot D and E are characterized by an equal distribution of almost all usage classes; (v) Hotspot F is mainly commercial and residential, with some institutional blocks; (vi) finally, Hotspot G presents all usage categories, with a prevalence of the commercial category. The higher the variety of the usage of a hotspot, the better a candidate this hotspot is for our analysis: then hotspots D and E are chosen. For these hotspots, and for 2015, Table \ref{tab:Result2015hotspotDE} shows the percentage of correctly classified patterns (among the normal and anomalous classes) obtained with 5 different trials in the form mean $\pm$ 95\% confidence. The classification performance is also calculated by using two well-known time series distance measures: the Dynamic Time Warping \cite{taylor:2015}, and the Fr\'echet distance \cite{frechet}. Clearly, the SRF measure outperforms both the DTW and the Fr\'echet distances.
\begin{table}[!t]
\renewcommand{\arraystretch}{1.3}
\caption{Percentage of Correct Classification achieved by analyzing of hotspot D and E during 2015, and using 3 similarity measures.}
\label{tab:Result2015hotspotDE}
\centering
% Some packages, such as MDW tools, offer better commands for making tables
% than the plain LaTeX2e tabular which is used here.
\begin{tabular}{|c|c|c|}
\hline
Similarity Measure  & 	Hotspot D      		&	Hotspot E			\\
\hline
SRF  				& 	95.61 $\pm$ 0.003 	& 	94.24 $\pm$ 0.24	\\
\hline
DTW  				& 	90.57 $\pm$ 0.134 	& 	91.80 $\pm$ 1.387	\\
\hline
FRECHET  			& 	90.14 $\pm$ 0.537 	& 	90.52 $\pm$ 2.34	\\
\hline
\end{tabular}
\end{table}
%}
In order to further test the assessment of anomalous patterns, 
the activity time series annotated as anomaly have been annotated by a triple according to their affinity with the typical pattern of each behavioral class. As an example, the triple W$|$E$|$L means that current time series is mostly similar to Working-Day typical pattern and secondly to Entertainment-Days one, whereas it shows only minor similarity with respect to Leisure-Days.
With the aim for measuring the capability of our similarity measure to generate a corresponding affinity assessment, we compute the average of the similarity of each time series annotated as an anomaly with respect to the most representative days of each behavioral class. Sorting them by similarity, we obtain the triple. The Mean Assessment Error is computed as the number of non-matching sort constraints for each pair of triples, averaged over all the set of the anomalies. As an example, the triples W$|$L$|$E and W$|$E$|$L have just one non-matching sort constraint, which is L$<$E, whereas both triples state that W$<$E and W$<$L.

A comparison is provided by repeating this procedure using the Dynamic Time Warping \cite{taylor:2015} distance. Resulting Mean Assessment Error are equal to 1.135 (SRF-based similarity measure) and 1.115 (DTW distance). According to these results both methods are suitable for pattern analysis, thus we provide the comparison of their performance in anomalous pattern detection.

Specifically, in order to compare the classification performances of our approach with respect to DTW, we collect the percentage of correctly classified days among 5 trials. During each trial, the DE generates a new set of Anomaly Index thresholds. If the Anomaly Index of an activity level time series exceeds the threshold, the corresponding day is considered anomalous. Obtained results are presented in the form “mean $\pm$ 95\% confidence interval” in Table \ref{tab:ClassificationResult}.

\begin{table}[!t]
\renewcommand{\arraystretch}{1.3}
% if using array.sty, it might be a good idea to tweak the value of
% \extrarowheight as needed to properly center the text within the cells
\caption{Percentage of Correct Classification achieved by analyzing data gathered during 2013, 2014, and 2015.}
\label{tab:ClassificationResult}
\centering
% Some packages, such as MDW tools, offer better commands for making tables
% than the plain LaTeX2e tabular which is used here.
\begin{tabular}{|c|c|c|}
\hline
Year  &        SRF        &        DTW      \\
\hline
2013  & 92.71 $\pm$ 0.321 & 90.57 $\pm$ 0.134\\
\hline
2014  & 96.65 $\pm$ 0.109 & 92.27 $\pm$ 0.106\\
\hline
2015  & 95.61 $\pm$ 0.003 & 91.28 $\pm$ 0.106\\
\hline
\end{tabular}
\end{table}

Based on obtained results, our approach provides an effective detection of major anomalies. But, handling minor or potential anomalies could be more difficult. In order to evaluate the effectiveness of our measure while handling this kind of anomalies, we select a set of events including official holidays and days affected by special events with documented effect on the road in (or in close proximity of the) hotspot D. Such events could be days characterized by adverse weather condition (e.g., Juno storm), street closure (e.g., due to the Gay Pride parade) and so on. This set is provided for each year under analysis. % and reported in Table 3...

\begin{figure}[!t]
\centering
\includegraphics[width=3in]{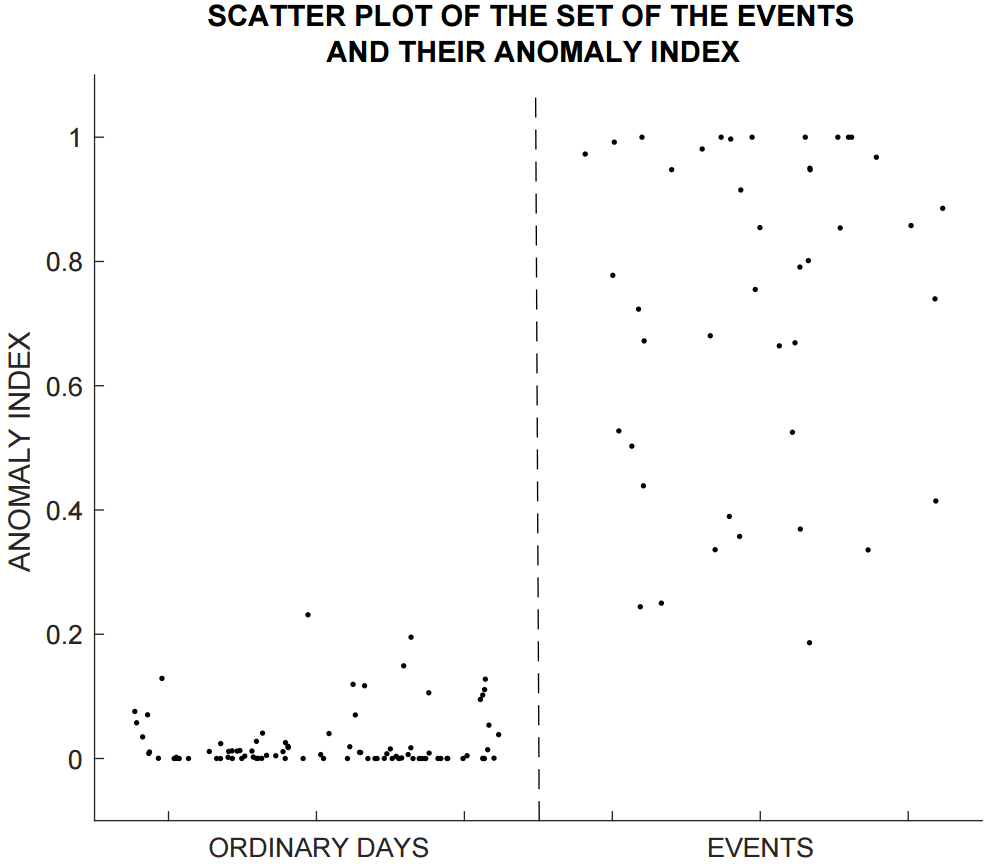}
\caption{Scatter-plot generated by considering Events, Typical Days and their corresponding Anomaly Index.}
\label{img:correlation}
\end{figure}

A set of ordinary days is also included. An effective anomaly measure is supposed to exhibit high correlation between its value and the set (events or ordinary days) which current day belongs to. In Fig. \ref{img:correlation} the correlation obtained by using our SRF-based measure is shown.

In order to compare obtained results in terms of correlation between events and computed Anomaly Index, we provide it by using SRF-based approach and DTW. In Table \ref{tab:correlation} we present obtained correlation coefficient for each year under analysis. 

\begin{table}[!t]
\renewcommand{\arraystretch}{1.3}
% if using array.sty, it might be a good idea to tweak the value of
% \extrarowheight as needed to properly center the text within the cells
\caption{Correlation coefficient between Anomaly Index and day characterized  by events occurring in the hotspot. Comparison between SRF-based approach and DTW.}
\label{tab:correlation}
\centering
% Some packages, such as MDW tools, offer better commands for making tables
% than the plain LaTeX2e tabular which is used here.
\begin{tabular}{|c|c|c|}
\hline
Year  &   SRF   &   DTW  \\
\hline
2013  & 0.8963  & 0.7177 \\
\hline
2014  & 0.9289  & 0.7236 \\
\hline
2015  & 0.9210  & 0.6828 \\
\hline
\end{tabular}
\end{table}

\section{Conclusion}
A novel approach for anomaly discovery in the context of taxi trip data have been proposed in this paper. 

Our approach was able to identify city hotspots, characterize the daily patterns of their activity over time and detect days characterized by an anomaly. Our approach has been tested on real world dataset containing all taxi trips occurred in Manhattan during 2013, 2014 and 2015, for a total amount of 74GB of data.   

On the basis of the results of this work, several conclusion have to be drawn. 

First of all, we prove the suitability of our approach for the analysis of "big data". This is due to the information abstraction provided by our multilayer topology.

Moreover, it is worth recalling that our approach exploits the information self-organization obtained by using the principle of stigmergy and does not require the in-depth modeling of the dynamics under investigation. This quality together with the adaptation provided by the SP training process, allows to specialize the detection on any occurred dynamics despite minor differences. Indeed, the performances of our approach have not been affected by differences among the typical activity patterns of each year.
The source code of our Stigmergic Perceptron has been publicly released on the MATLAB Central File Exchange \cite{sourceCode}.

The performances of our approach is measured in terms of percentage of correctly classified daily patterns among typical and anomalous ones. Moreover, the effectiveness in handling minor anomalies is measured by computing the correlation between the Anomaly Index and the occurrence of urban events (e.g., local parade or official holiday). 

Obtained results have been compared with respect to the one achieved by analyzing the activity time series with DTW. In both cases, and in every year under analysis, our approach outperforms DTW, achieving up to 96.65 percentage of correctly classified days and 0.9289 correlation coefficient. 

Although mobility in Manhattan can be characterized by many dynamics for each day, both typical and exceptional, the stigmergy-based approach has proven to be a withstanding method to enhance significant patterns in taxi trip data, letting them emerge autonomously. 

Finally yet importantly, our approach provides a continuous measure (the Anomaly Index) of the divergence with respect to typical activity patterns, which can be used by policy makers to evaluate the effect and the resilience of proposed policies or change them dynamically according to provided measures.
%\textcolor{blue}{
Our approach can be potentially applied to any sustainable transportation indicator, to distinguish different spatiotemporal patterns occurring over time. For each indicator and for each basic pattern, a subset of labelled data is required to enable a proper calibration and interpretation of what can be discovered. This study focuses on a complete description, experimentation and discussion of our approach via a specific indicator extracted from a real world taxi trips data set. A more extensive experimentation on multiple indicators is clearly possible but it is out of the scope of this work. 
%}
Thus, one of the most promising improvements for this study consists of cross-checking results obtained by exploiting different data sources and indicators. As an example, by incorporating pollution data or car crash data into the analysis, policy makers could gain insight about the mutual interaction of this phenomena, which is a fundamental knowledge for producing effective policies.

\ifCLASSOPTIONcaptionsoff
  \newpage
\fi



%\bibliographystyle{IEEEtran}

\begin{IEEEbiography}[{\includegraphics[width=1in,height=1.25in,clip,keepaspectratio]{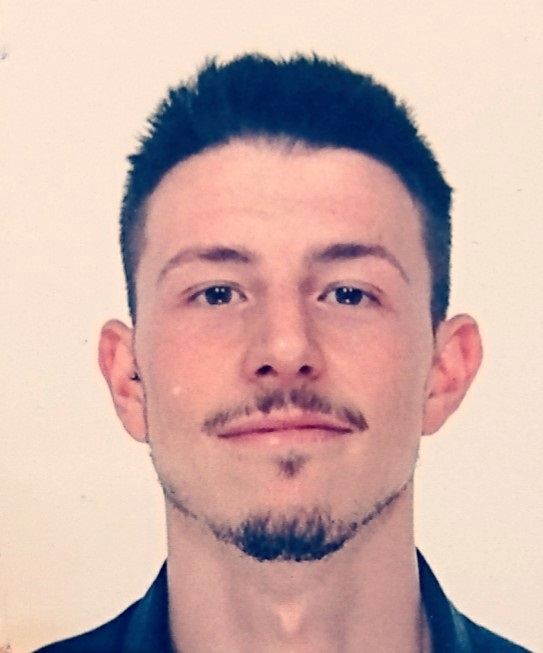}}]{Antonio Luca Alfeo}
%\begin{IEEEbiography}{Antonio Luca Alfeo}

was born in Taranto, Italy on September 4, 1987. He received the B.S. and 
the M.S. degree in Computer Engineering from the University of Pisa, in 2012 and 2015 respectively.
He is currently a Ph.D. student of the International Ph.D. Program in Smart Computing and a visiting Ph.D. student at the Media Lab of the Massachusetts Institute of Technology. 
His research interests include the design and development of emergent and bio-inspired approaches for machine learning and data analysis, with applications in the context of cyber-physical systems including collaborative UAVs swarms, e-health, and smart cities.
\end{IEEEbiography}

\begin{IEEEbiography}[{\includegraphics[width=1in,height=1.25in,clip,keepaspectratio]{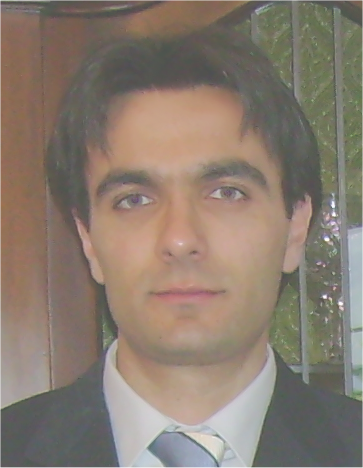}}]{Mario G.C.A. Cimino}
is with the Department of Information Engineering (University of Pisa) as an Associate Professor. He is also a research associate at the Institute for Informatics and Telematics (IIT) of the Italian National Research Agency (CNR). In 2006, he was a visiting Ph.D. student at the Electrical and Computer Engineering Research Facility of the University of Alberta, Canada. In 2007, he received the Ph.D. degree in Information Engineering from the University of Pisa. His research focus lies in the areas of Swarm Intelligence and Business/Social Process Analysis, with particular emphasis on Stigmergic Computing, Workflow Mining and Simulation. He is (co-) author of more than 50 publications.
\end{IEEEbiography}

\begin{IEEEbiography}[{\includegraphics[width=1in,height=1.25in,clip,keepaspectratio]{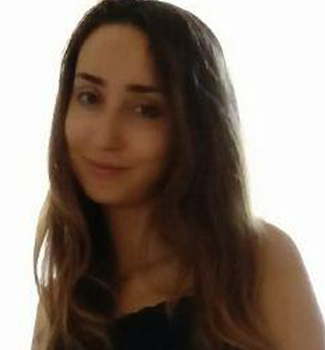}}]{Sara Egidi}
was born in Ascoli Piceno, Italy, on 18 November 1990.
She received the B.S. degree in industrial computer science from the University of Camerino, Ascoli Piceno, Italy, in 2014 and the M.S. degree in computer engineering from University of Pisa, Pisa, Italy, in 2017. Her research has been concerned with machine learning techniques applied in the field of city science.
\end{IEEEbiography}

\begin{IEEEbiography}[{\includegraphics[width=1in,height=1.25in,clip,keepaspectratio]{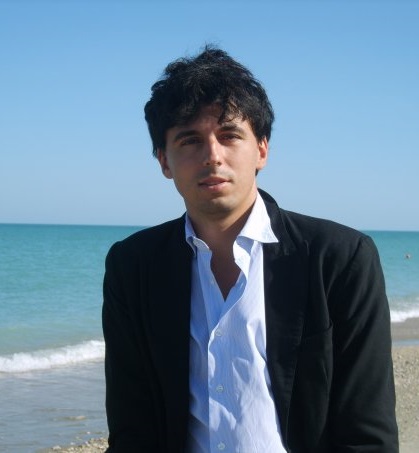}}]{Bruno Lepri}
Bruno Lepri leads the Mobile and Social Computing Lab (MobS) and is vice-responsible of the Complex Data Analytics research line at Bruno Kessler Foundation (Trento, Italy). Bruno is also research affiliate at the MIT Connection Science initiative and he has recently launched an alliance between MIT Connection Science and Bruno Kessler Foundation on Human Dynamics Observatories. He is also a senior research affiliate of Data-Pop Alliance, the first think-tank on Big Data and Development co-created by the Harvard Humanitarian Initiative, MIT Media Lab, Overseas Development Institute, and Flowminder to promote a people-centered big data revolution. In 2010 he won a Marie Curie Cofund post-doc fellowship and he has held post-doc positions at FBK and at MIT Media Lab. He holds a Ph.D. in Computer Science from the University of Trento. His research interests include computational social science, personality computing, urban computing, network science, machine learning, and new models for personal data management and monetization. His research has received attention from several international press outlets and obtained the James Chen Annual Award for best UMUAI paper and the best paper award at ACM Ubicomp 2014. His work on personal data management was one of the case studies discussed at the World Economic Forum.
\end{IEEEbiography}

\begin{IEEEbiography}[{\includegraphics[width=1in,height=1.25in,clip,keepaspectratio]{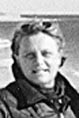}}]{Gigliola Vaglini}
 was born in Pisa, Italy, on June 11, 1952. She received the M.S. degree in Computer Science from the University of Pisa. She was a research assistant at the University of Pisa, Department of Computer Science, an associate Professor at the University of Naples, Federico II, Department of Mathematics, and then at the University of Pisa, Department of Information Engineering; from 2002 she is a Full Professor at the same Department. Her research has been concerned mainly with formal methods for specification and verification of concurrent and distributed systems, in particular the model checking of concurrent systems. More recently her research activity was focused on the use of the stigmergic paradigma to aggregate data supplied by different sources and to obtain a distributed control on autonomous systems.
\end{IEEEbiography}


%\newpage



%\vfill


\end{document}